\documentclass[sigconf]{acmart}

\renewcommand\footnotetextcopyrightpermission[1]{} 
\usepackage{amsmath,graphicx}
\usepackage{caption}
\usepackage{subcaption}

\AtBeginDocument{%
  \providecommand\BibTeX{{%
    \normalfont B\kern-0.5em{\scshape i\kern-0.25em b}\kern-0.8em\TeX}}}

\begin{document}

\title{An Integrated System for Mobile \\ Image-Based Dietary Assessment}


\author{Zeman Shao$^{*}$, Yue Han$^{*}$, Jiangpeng He$^{*}$, Runyu Mao$^{*}$}
\author{Janine Wright$^{\dagger}$, Deborah Kerr$^{\dagger}$, Carol Boushey$^{\ddagger}$, Fengqing Zhu$^{*}$}
\affiliation{
  \institution{$^{*}$School of Electrical and Computer Engineering, Purdue University, West Lafayette, Indiana, USA}
  \institution{$^{\dagger}$School of Public Health, Curtin University, Perth, Western Australia}
  \institution{$^{\ddagger}$Cancer Epidemiology Program, University of Hawaii Cancer Center, Honolulu, Hawaii, USA}
  \country{}
}

\renewcommand{\shortauthors}{}

\begin{abstract}
Accurate assessment of dietary intake requires improved tools to overcome limitations of current methods including user burden and measurement error. Emerging technologies such as image-based approaches using advanced machine learning techniques coupled with widely available mobile devices present new opportunities to improve the accuracy of dietary assessment that is cost-effective, convenient and timely. However, the quality and quantity of datasets are essential for achieving good performance for automated image analysis. Building a large image dataset with high quality groundtruth annotation is a challenging problem, especially for food images as the associated nutrition information needs to be provided or verified by trained dietitians with domain knowledge. In this paper, we present the design and development of a mobile, image-based dietary assessment system to capture and analyze dietary intake, which has been deployed in both controlled-feeding and community-dwelling dietary studies. Our system is capable of collecting high quality food images in naturalistic settings and provides groundtruth annotations for developing new computational approaches.
\end{abstract}





\maketitle
\pagestyle{plain}

\section{Introduction}
\label{sec:intro}
Health conditions linked to poor diet constitute the most frequent and preventable causes of death in the United States and are major drivers of health care costs, estimated in the hundreds of billions of dollars annually~\cite{cdc-cost}.
Despite this strong association, our ability to accurately assess dietary intake remains a challenge in nutrition science. 
Traditional dietary assessment is comprised of written and orally reported methods~\cite{Thompson2017} which has major limitations due to participant burden and recall or memory bias~\cite{Poslusna2009,livingstone2004}. 
Emerging mobile and wearable technologies provide unique mechanisms for collecting food images in context, showing great potentials for advancing the field of dietary assessment.

Recently, deep learning based approaches have been developed for image-based dietary assessment methods to analyze food images including food recognition and portion estimation~\cite{Meyers_2015_ICCV,yanai_2017,he2020multi}.
Most deep learning methods are data-hungry and their performances are heavily reliant on the quantity and quality of the available datasets.
However, building a large image dataset with high quality groundtruth labels in a systematic way is challenging.
Annotation tools and platforms, such as LabelMe~\cite{russell2008labelme} and Amazon Mechanical Turk (AMT)~\cite{amtwebsite}, are widely used to perform image annotation such as creating bounding boxes to indicate the pixel locations and label of the objects within.
In~\cite{rabbi_15}, a machine learning method is proposed to identify high performing worker on AMT in order to achieve high accuracy.
In~\cite{shao2019semi}, a deep neural network is developed for food object detection to remove noisy images which do not contain foods during pre-processing, alleviating the burden placed on human annotators.
However, none of the existing annotation tools and platforms provide systematic ways to collect food images and to obtain accurate dietary information which requires trained dietitians with domain knowledge.

Previously, we designed and developed the Technology Assisted Dietary Assessment (TADA) system that records and analyzes images of eating occasions~\cite{zhu2010A,six2010}. The TADA system consists of three main components: the mobile Food Record (mFR), the secure cloud based server, and the web interface. 
The mFR is a mobile app that uses the camera on a smartphone to capture images of food consumed during an eating occasion, and allows participants to review and confirm location and labels of foods in the image resulted from image analysis~\cite{tada_ziad}. 
The mFR sends food images, along with metadata (\textit{e.g.}, date, time, GPS coordinates) to a cloud based server where the food images are segmented, the food types are classified, and portion sizes are estimated using image analysis, computer vision, and machine learning methods~\cite{tada_segmentation,wang-icip2017,Mao2021,Fang2018}. 
The web interface is designed for nutrition researchers to visualize and verify collected food images and the associated dietary information, supported by an integrated database system~\cite{Bosch2011a}.
In this paper, we enhance various components of the TADA system by proposing a systematic design to collect food images from study participants and to provide accurate dietary information that is meaningful to nutrition researchers. 

The contributions of this paper are twofold.
First, we propose a system which provides an efficient mechanism for image-based dietary assessment to support nutrition science researchers.
Second, the proposed system is capable of building a comprehensive food image dataset with high quality visual and nutrition groundtruth for developing novel computational approaches.

\section{System Design}
\label{sec:overview}
\begin{figure}[t!]
	{\includegraphics[width =8.5cm]{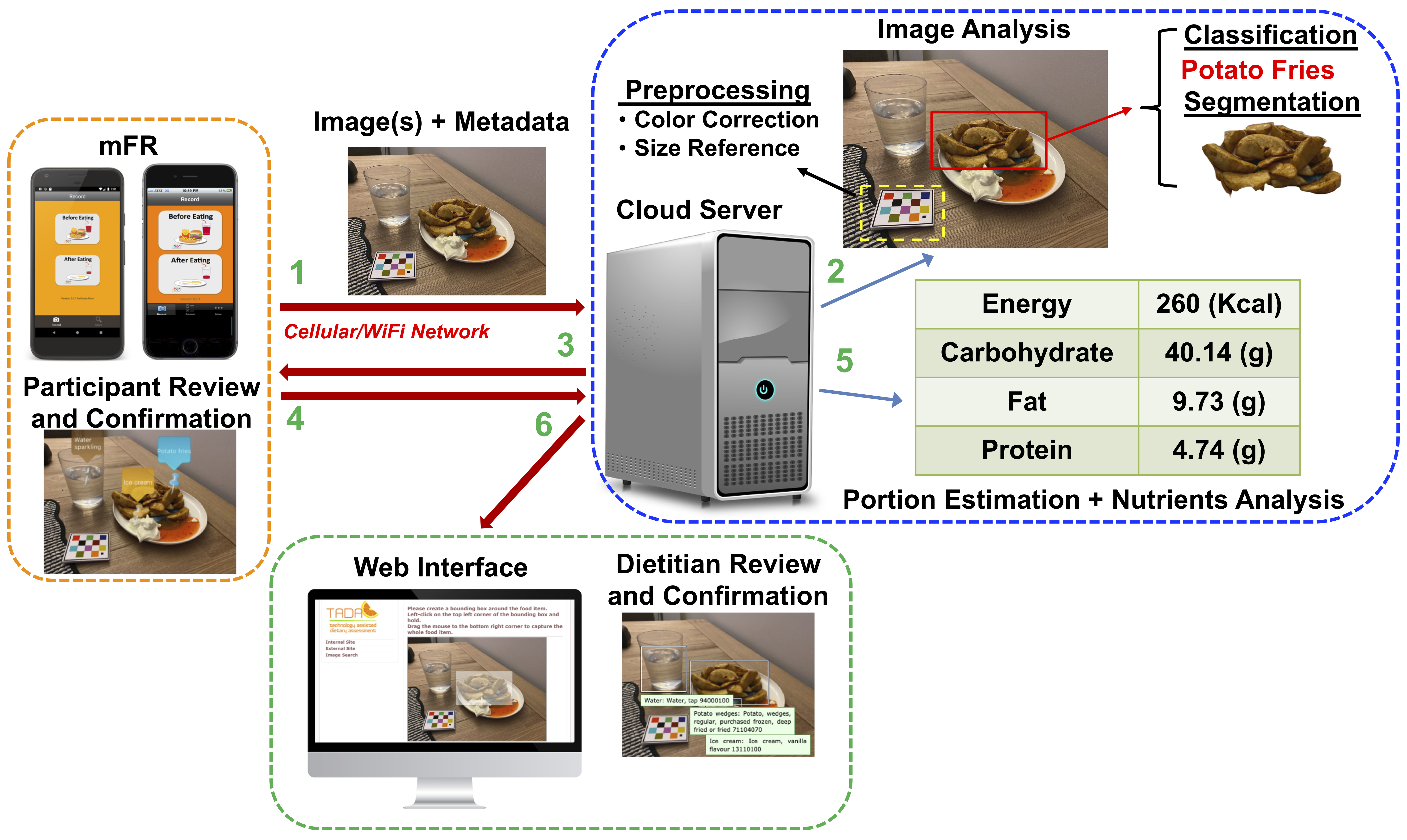}\label{fig:overview}}
	\hfill
	\caption{Overview of the enhanced TADA system.}
	\label{fig:overview}
\vspace{-1.0em}
\end{figure}

An overview of the enhanced TADA system is shown in Figure~\ref{fig:overview}. 
The mobile Food Record (mFR) captures a pair of before and after eating images. 
These images are sent to the cloud server along with metadata including date, time, GPS coordinates, camera pose angle and image Exif data (Step 1). 
Image analysis is done on the cloud server (Step 2) and predicted food labels and corresponding pixel locations are sent back to the mFR (Step 3). 
The mFR provides interface for the participant to review, modify and confirm the food labels received from the server (Step 4). 
Participant confirmed results are then sent back to the server for additional refinement and portion size estimation (Step 5). Results can then be viewed and interacted with via the web interface by nutrition researchers from any computer with an internet connection to perform nutrition analysis (Step 6). 

\subsection{Mobile Food Record}
\begin{figure}
     \centering
     \begin{subfigure}[b]{0.23\textwidth}
         \centering
         \includegraphics[width=\textwidth]{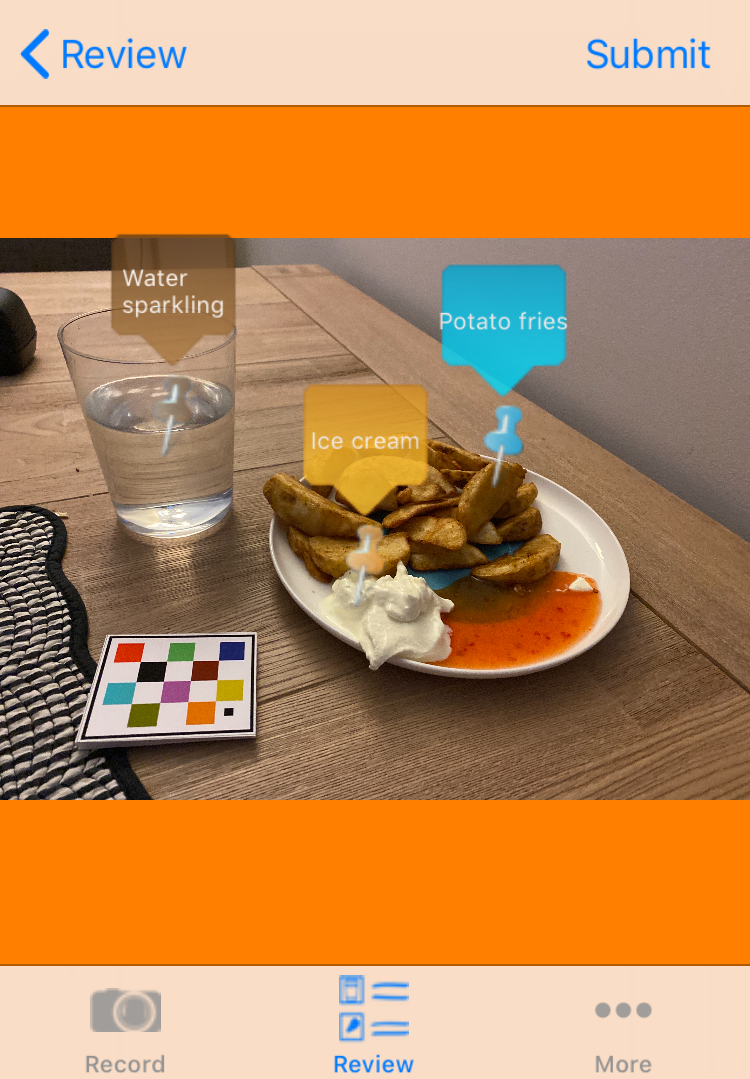}
         \caption{Review and modification}
         \label{fig:mfr_annotation}
     \end{subfigure}
     \hfill
     \begin{subfigure}[b]{0.23\textwidth}
         \centering
         \includegraphics[width=\textwidth]{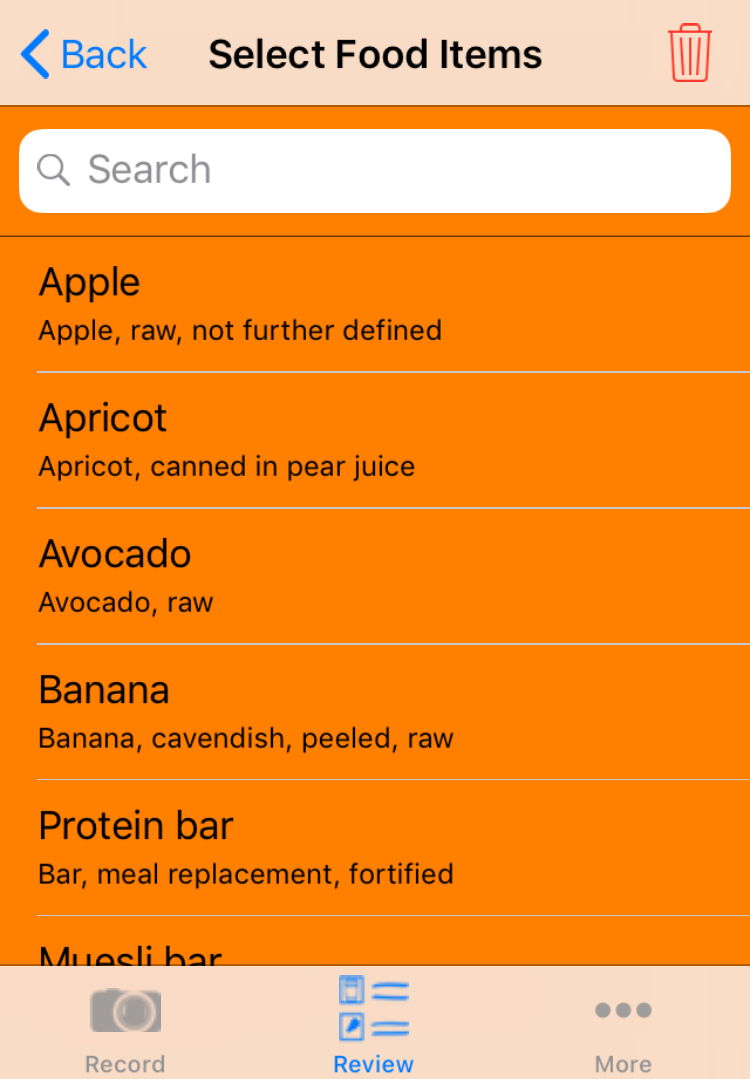}
         \caption{Food list with description}
         \label{fig:mfr_list}
     \end{subfigure}
    \caption{mFR review process.}
    \label{fig:mfr}
\vspace{-1.0em}
\end{figure}

The Mobile Food Record (mFR) allows participant to record the eating occasion images and review the image analysis results from the server. Participants record the foods and beverages consumed at a meal by capturing a pair of images, one image before eating and one image after eating. This pair of images are sent to the server for analysis. Results are sent back to the mFR for participants to review, modify, and confirm.  This information can be used by the image analysis done on the server to update the learning methods. The review process requires participant to confirm or modify the food labels by clicking the pins on the before image (Figure~\ref{fig:mfr_annotation}). A list of foods relevant to the nutrition study is pre-loaded on the mFR to allow participants to choose from (Figure~\ref{fig:mfr_list}). Details on the mFR interface design and functionalities can be found in~\cite{tada_ziad}.

\subsection{Cloud Server}
The cloud server is responsible for the data transmission, storage and inference.
The server receives the eating occasion images and associated metadata from the mFR. This information is stored in an integrated database systems~\cite{Bosch2011a}. 
The cloud server provides preliminary image analysis results that are sent back to the mFR for the participant to review and confirm. Results are further refined, from which portion estimation and nutrition analysis are performed. 
Image analysis and inference include food segmentation, food classification and portion size estimation~\cite{wang-icip2017,Mao2021,Fang2018}.
This information can be visualized and used for additional nutrition analysis using the web interface by nutrition researchers.

\subsection{Web Interface}
\label{sec:web_interface}
\begin{figure}[t!]
	\centering
	{\includegraphics[width=7 cm]{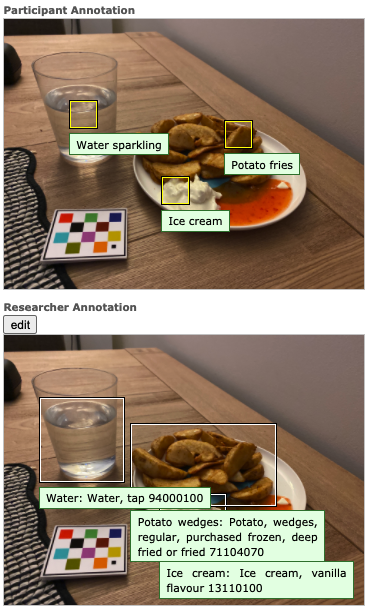}\label{fig:web_preview}}
	\hfill
	\caption{Participant confirmed preliminary results and researcher confirmed final annotations.}
	\label{fig:web_preview}
\vspace{-1.0em}
\end{figure}

In addition to visualizing the collected data from dietary studies, we designed a set of new features for the web interface and incorporated them into the TADA system. 
Our goal is to design web-based annotation tools that can be interacted by nutrition researchers to review and modify participant confirmed results and to conduct additional nutrition analysis. 
The annotation tool provides a list of foods with associated food codes from the USDA Food and Nutrition Database for Dietary Studies (FNDDS) database~\cite{fndds2018} and a searching tool to quickly return relevant food labels. 
Our design takes into consideration the following criteria including easy and quick accessibility for researchers to modify the preliminary results from the participants, and high quality annotation results that can be used to improve the accuracy of the image analysis.

The web-based annotation tools consist of two key ``steps" for researchers to access and modify participant confirmed preliminary results. 
These steps include review of collected food image and its associated dietary information, and the modification of participant confirmed preliminary results.  

\textbf{Review of collected food image and associated dietary information.} 
For each participant, all captured before and after eating scene images are displayed as small preview images on the web interface. 
Researchers can click either the before or after eating scene preview image to review images captured for that eating occasion, along with participant confirmed food types and researcher confirmed final annotations as shown in Figure~\ref{fig:web_preview}.
The annotations from participants and researchers are displayed on the eating scene image with the food labels and bounding boxes information. 
An example is shown in Figure~\ref{fig:web_preview} where some foods are not labeled correctly by the participants and the bounding box is not accurately.
The web interface provides an interactive design to modify the participant confirmed preliminary results.

\textbf{Modification of participant confirmed preliminary results.} 
The web-based annotation tool has a dedicated view for the researcher to modify the annotations received from participants. This function can be accessed by clicking the ``edit" button below the researcher annotation as shown Figure~\ref{fig:web_preview}).
The researcher annotation process consists of three main operations: entering the initials of the researcher performing the task, drawing a bounding box around the food item in the image, selecting the corresponding food label to save the bounding box. The process can be repeated if there are multiple foods in the eating scene image. 

\begin{figure}
     \centering
     \begin{subfigure}[b]{0.48\textwidth}
         \centering
         \includegraphics[width=\textwidth]{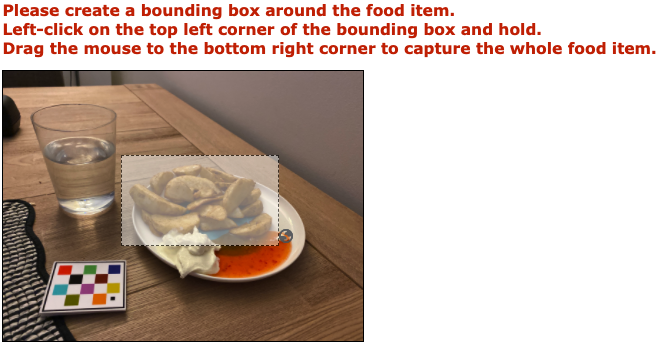}
         \caption{The interface for drawing a bounding box.}
         \label{fig:web_annotation}
     \end{subfigure}
     \hfill
     \begin{subfigure}[b]{0.48\textwidth}
         \centering
         \includegraphics[width=\textwidth]{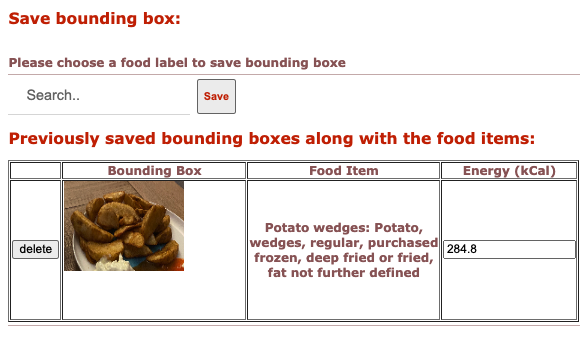}
         \caption{Associated information for the selected bounding box.}
         \label{fig:web_confirmation}
     \end{subfigure}
    \caption{Examples of the web-based annotation tools.}
    \label{fig:web_tools}
\vspace{-1.0em}
\end{figure}

As shown in Figure~\ref{fig:web_annotation}, researchers can modify or add a food by drawing a bounding box on the eating scene image using the hold-and-drag operation common to all operating systems. 
The instruction for this step is displayed on the top of the web interface to provide clear guideline to the researchers. Once the bounding box is drawn, a cropped image of the extract region is shown in the review section.  This allows researchers to verify the bounding box is accurately draw, particularly for more complex eating scenes with multiple food items.

The researcher can select a food label to match the cropped region of the image by using the search bar on the web interface, which contains a pre-loaded food list and associated USDA FNDDS food codes. 
The search can be done by typing in either the food name or the food code in the search bar. The search mechanism is similar to what is used in most operating systems which shows all relevant results containing the keywords. 
For example, when `potato' is typed, then `potato', `potato wedges' and `roast potato' are shown in search bar. Because some food names are the same, the matched food codes for the food names are shown to differentiate them. If the food label is not on the pre-loaded food list, the researcher can click the save button directly and the keyword entered will be saved as the food label for the bounding box. Once the food label is selected for the bounding box, corresponding energy information can be entered manually by the researcher. An example of the saved food region, matched food name and nutrition information are shown in Figure ~\ref{fig:web_confirmation}). A delete option is provided if there is any mistake made during this process. Once the research can click the Finish button at the bottom of the web page once information for all foods in an image is completed. The aggregated information from the researcher and the participant is made available for download.   

\section{Deployment in Dietary Studies}
\label{sec:study}
One main goal of our system is to provide nutrition researchers an efficient mechanism to collect high quality eating occasion images and nutrition information for dietary assessment.
The system has been widely deployed in more than 30 dietary studies with over 2,500 participants between the ages 6 months – 70 years in domestic and international locations.  We have collected more than 72,000 images from both controlled feeding and community dwelling studies. Table~\ref{table:study} describes some of the published studies.
Our system has shown to be acceptable and feasible for dietary studies as indicated in \cite{boushey2017reported} where participants were asked to capture images of all eating occasions over 7.5 days. 
Prior to using the mFR, 71\% of them agreed “Remembering to take an image before meals would be easy.” 
After using the mFR for 7.5 days, the agreement rate reached 100\%.
For capturing food images, participants were provided with a known-dimension fiducial marker placed in eating scene for providing physical and color references. 
Initially 87\% thought it would be easy to use and these responses changed to 96\% after 7.5 days.

\begin{table}[h]
\begin{center}
 \begin{tabular}{|c c c c|} 
 \hline
 Study & Location & Population & Age \\
 \hline\hline
 \cite{aflague2015feasibility}     & Guam                  & 65            & 3-10 years    \\ 
 \hline
 \cite{bathgate2017feasibility}     & Perth, West Australia & 58            & 12-30 years    \\ 
 \hline
 \cite{boushey2017reported}     & Tippecanoe County, IN & 45            & 21-65 years    \\ 
 \hline
 \cite{boushey2016dietary}     & Pacific Coast, Washington & 24        & 21-60 years    \\ 
 \hline
 \cite{daugherty2012novel}     & Tippecanoe County, IN & 57            & 21-65 years    \\ 
 \hline
 \cite{kerr2016connecting}     & Perth, West Australia & 247           & 18-30 years    \\ 
 \hline
 \cite{halse2019improving}     & Perth, West Australia & 165           & 18-65 years    \\ 
 \hline
 \cite{panizza2017barriers}     & O'ahu                 & 93            & 9-13 years    \\ 
 \hline
 \cite{panizza2019effects}     & O'ahu                 & 60            & 35-55 years    \\ 
 \hline
 \cite{polfuss2018technology}     & Milwaukee, WI         & 12            & 8-18 years     \\ 
 \hline
 \cite{six2009evaluation}     & Tippecanoe County, IN & 63            & 11-18 years     \\ 
 \hline
 \cite{solah2017effect}     & Perth, West Australia & 118           & 25-70 years    \\ 
 \hline
\end{tabular}
\end{center}
\caption{Summary of published studies using the system.}
\label{table:study}
\vspace{-1.0em}
\end{table}
\vspace{-1.0em}

\section{Food Image Analysis}
\label{sec:result}

Accurate estimation of dietary intake relies on the system’s ability to distinguish foods from image background (\textit{i.e.}, localization and segmentation), to identify or label the foods (\textit{i.e.}, classification), to estimate food portion size, and to understand the context of the eating event. We have developed vision-based solutions to many challenging aspects of novel image-based dietary assessment tools. On food recognition, we developed a two-step approach for food localization and hierarchical food classification using Convolutional Neural Networks (CNNs) to reduce prediction error for visually similar foods \cite{Mao2021}, and continual learning in the challenging online learning scenario that is further bounded by run-time and limited data \cite{he-cvpr2020,he2021online}. On image segmentation, we developed class-agnostic method using a pair of eating scene images to find the salient missing objects without prior information about the food class \cite{tada-yarlagadda-acm} and developed weakly supervised and efficient superpixel based methods \cite{wang-icip2017,wang2016efficient}. On food portion size estimation, we developed different approaches using a single-view image, including the use of geometric models to recover 3D parameters of food objects in the scene \cite{fang-ism2015}, incorporating co-occurrence patterns to refine portion estimation results \cite{fang_2017}, and deep learning for the mapping of food images to food energy \cite{Fang2018,fang2019end}. To understand the contextual and dynamic attributes of diet, we explored various aspects of eating context and environment that influence dietary intake, such as combining temporal information and food co-occurrence to develop personalized learning model \cite{Wang2017a}, and understanding different eating environments through novel image clustering \cite{Yarlagadda2020}.  



\begin{figure}
    \centering
    \includegraphics[width=0.95\linewidth]{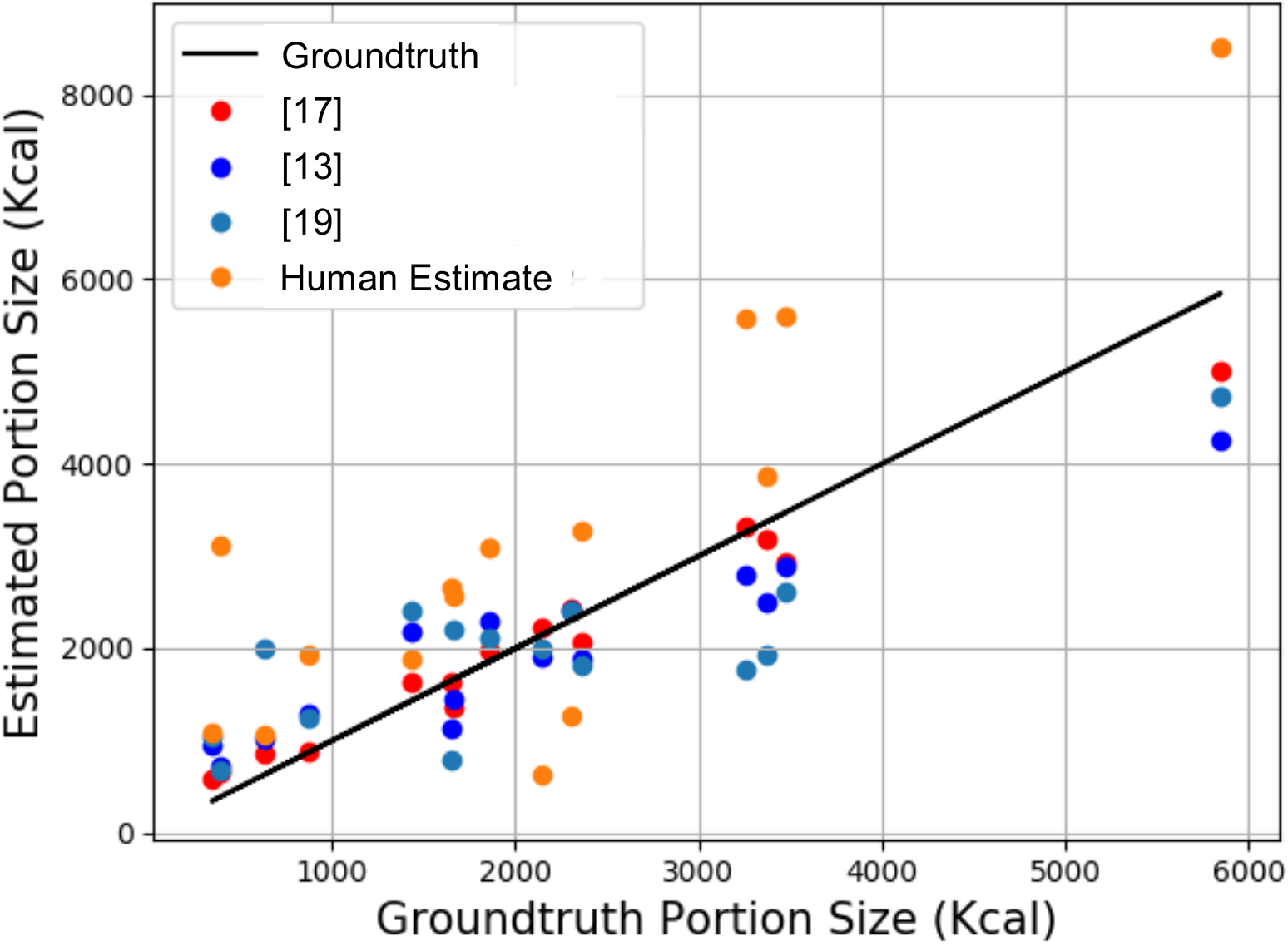}
    \caption{Relationship between groundtruth and estimated food energy by human and deep learning models. The black line indicates the groundtruth and estimated energy are the same where the dots above this line indicate overestimation, and the dots below indicates underestimation. Different colored represent results from using different methods~\cite{fang2019end,he2020multi,he2021end} for comparison. (Best viewed in color)}
    \label{fig:portion_results}
    \vspace{-1.0em}
\end{figure}

In addition to developing methods for individual image analysis tasks, We have also developed integrated food image analysis systems including multiple hypothesis approach \cite{tada_segmentation}, and deep learning based methods that jointly performs food localization, classification, and portion size estimation \cite{he2020multi,he2021end}. Figure~\ref{fig:portion_results} shows results of improved portion estimation from different approaches. We observed that estimating food energy accurately from an eating scene image is a challenging task for people who do not have domain knowledge, note that some people were not able to recall items they consumed. Trained on data collected by our proposed system in a dietary study \cite{feedingstudy}, our best method achieved a mean error rate of 11.22\%, which significantly outperforms the human estimation error of 62.14\%.

\section{Conclusion}
We enhanced a previously proposed image-based dietary assessment system by implementing new functionalities to efficiently collect food images and generate high quality visual and nutrition groundtruth information in a systemic design. Our system provides the necessary tools for researchers to conduct both image analysis and nutrition research more efficiently. We envision this system would open doors to many possible image-based dietary assessment applications using deep learning techniques. Our integrated system holds promise to improving the system efficiency and accuracy, reducing both participant and researcher burden associated with traditional approaches.
\bibliographystyle{ACM-Reference-Format}
\bibliography{main}

\end{document}